\documentclass[letterpaper, 10 pt, journal, twoside]{IEEEtran}
\ifCLASSINFOpdf
\else
\fi

\usepackage{graphicx}
\usepackage{multirow}
\usepackage{amsmath}
\usepackage{amssymb}
\usepackage{array}
\usepackage[font={small}]{caption}
\usepackage{subcaption}
\usepackage[dvipsnames]{xcolor}
\usepackage{color,soul}
\usepackage{url}
\usepackage[ruled,vlined]{algorithm2e}
\usepackage{booktabs}
\usepackage{bm}

\begin{document}
%
\title{Distributional Reinforcement Learning based Integrated Decision Making and Control for Autonomous Surface Vehicles}
%
%
%

\author{Xi Lin, Paul Szenher, Yewei Huang, and Brendan Englot%
\thanks{This work was supported by the Office of Naval Research, Grants N00014-20-1-2570 and N00014-24-1-2522.} 
\thanks{The authors are with the Department of Mechanical Engineering, Stevens Institute of Technology, Hoboken, NJ, USA
{\tt\footnotesize \{xlin26,pszenher,yhuang85,benglot\}@stevens.edu}}
}
\maketitle

\begin{abstract}
With the growing demands for Autonomous Surface Vehicles (ASVs) in recent years, the number of ASVs being deployed for various maritime missions is expected to increase rapidly in the near future.
However, it is still challenging for ASVs to perform sensor-based autonomous navigation in obstacle-filled and congested waterways, where perception errors, closely gathered vehicles and limited maneuvering space near buoys may cause difficulties in following the Convention on the International Regulations for Preventing Collisions at Sea (COLREGs).
To address these issues, we propose a novel Distributional Reinforcement Learning based navigation system that can work with onboard LiDAR and odometry sensors to generate arbitrary thrust commands in continuous action space.
Comprehensive evaluations of the proposed system in high-fidelity Gazebo simulations show its ability to decide whether to follow COLREGs or take other beneficial actions based on the scenarios encountered, offering superior performance in navigation safety and efficiency compared to systems using state-of-the-art Distributional RL, non-Distributional RL and classical methods.
\end{abstract}


%
\IEEEpeerreviewmaketitle

\vspace{-3mm}
\section{Introduction}
\vspace{-1mm}
Recent years have witnessed increasing demands for and deployments of Autonomous Surface Vehicles (ASVs), which motivates the development of autonomous navigation systems \cite{vagale2021path}.
This paper considers scenarios in which multiple ASVs operate in congested waterways where buoys may exist, and aims at providing a single robust onboard ASV navigation system which can be deployed on multiple ASVs.   
Some existing works have considered collision avoidance for multiple vessels with simplified vehicle and environment dynamics, and assumed very accurate perception in simulation \cite{zhao2019colregs,cho2020efficient}.
Other works have verified algorithm performance in field experiments, but with only one autonomous vehicle and other vessels driven by humans or following pre-defined trajectories \cite{kuwata2013safe,hagen2018mpc,eriksen2019branching}, without demonstrating performance in multi-ASV navigation scenarios. 

\begin{figure}
    \centering
    \includegraphics[width=0.99\linewidth]{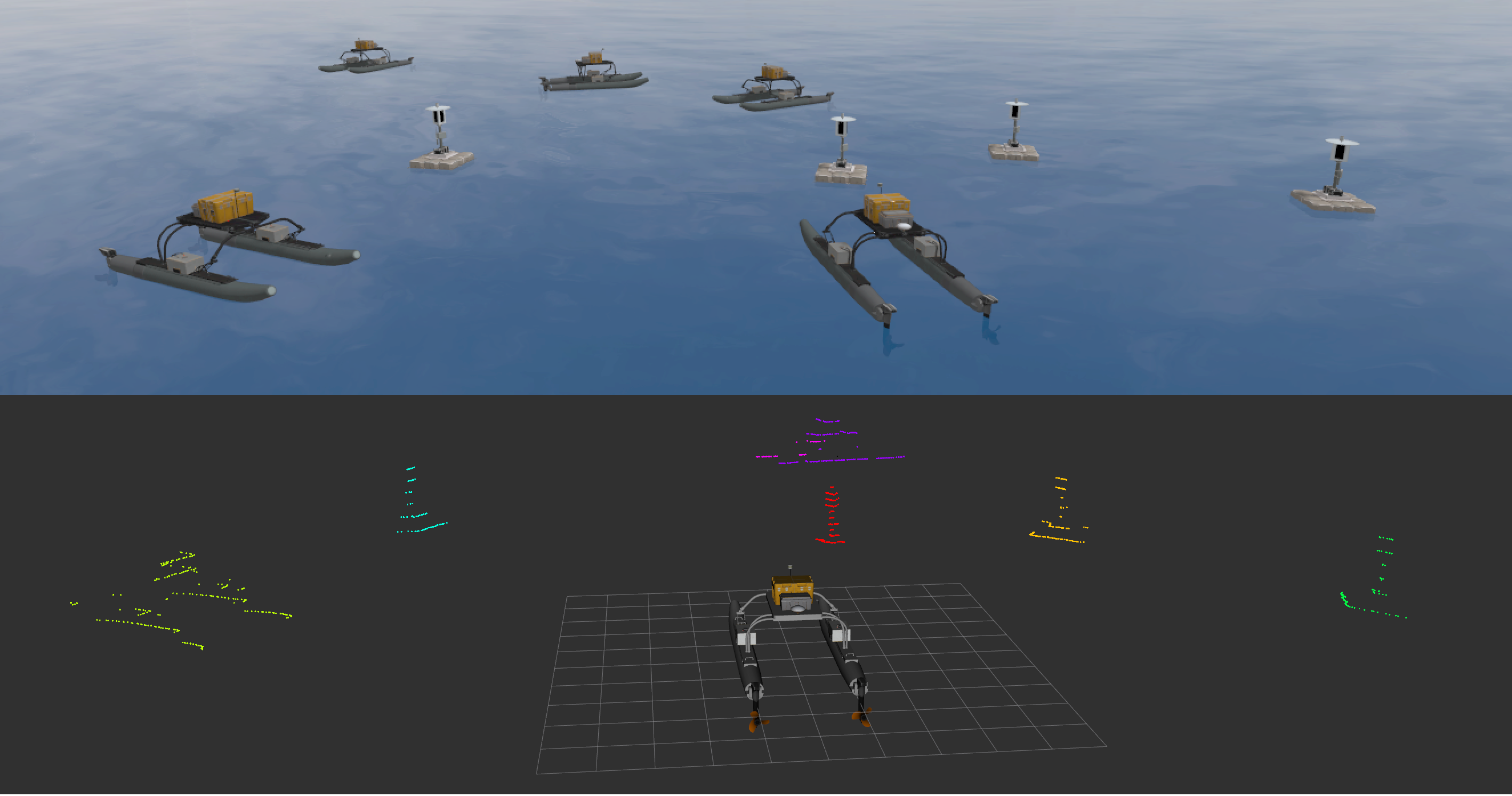}
    
    \vspace{-0.1mm}
    \includegraphics[width=0.99\linewidth]{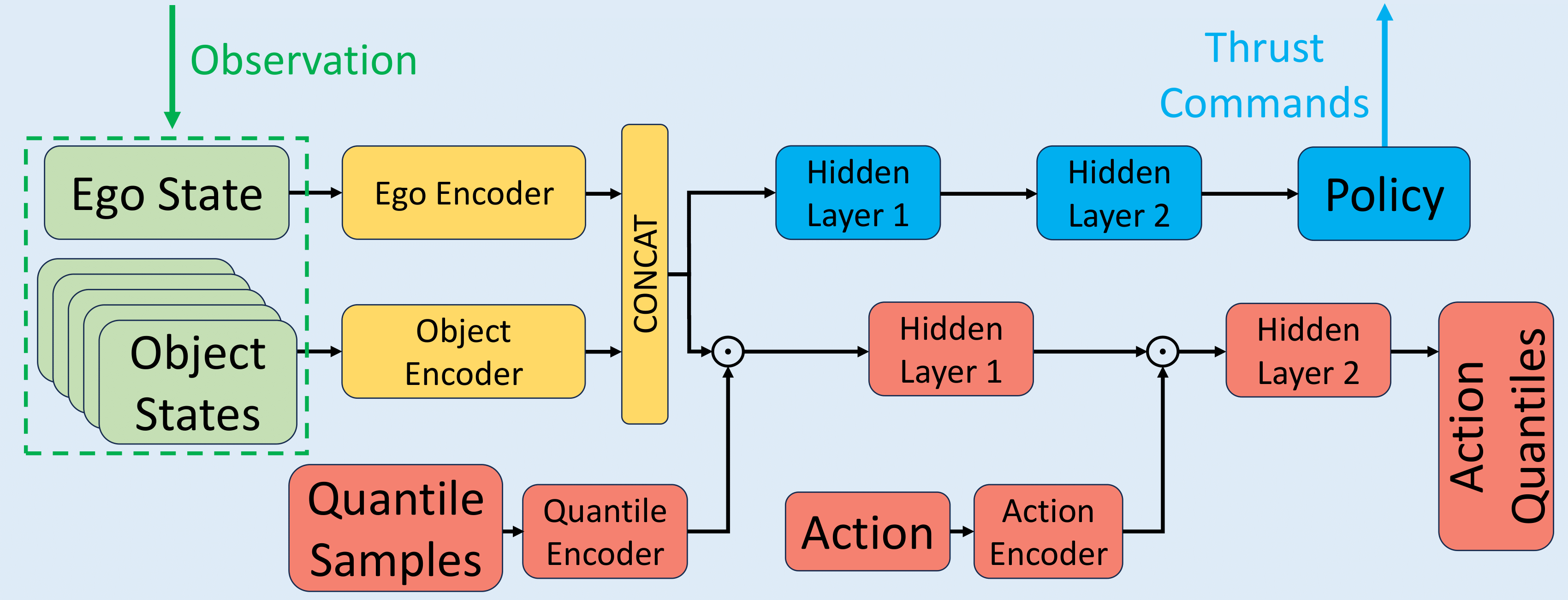}
    \caption{{\bf The proposed navigation system}. 
    The top figure shows the perspective in the Gazebo simulation. 
    The middle figure visualizes the segmentation result of LiDAR point clouds received by the lower right vehicle in the top figure.
    The bottom figure illustrates the decision making and control module of the proposed system. 
    }
    \label{fig:system}
    \vspace{-6mm}
\end{figure}

In this letter, we propose a novel and computationally efficient ASV navigation system that can work with onboard LiDAR and odometry sensors, and we demonstrate its performance in a Gazebo based simulator \cite{VRX} that is capable of simulating complex multi-ASV navigation scenarios with realistic environment dynamics, vehicle models, and perceptual sensing.

The framework of our proposed system is shown in Figure \ref{fig:system}, which uses a Distributional Reinforcement Learning (Distributional RL) agent for decision making and control of the vehicle given perceptual information.
The Distributional RL agent can be represented as a deep neural network model, and learns the distributions of cumulative rewards related to different actions via trial-and-error learning \cite{bellemare2023distributional}.
The robustness of Distributional RL policies in the multi-ASV navigation problem has been demonstrated in our prior work \cite{lin2024decentralized}, which uses Implicit Quantile Networks (IQN) \cite{dabney2018implicit}. 

However, the limitations of our prior work \cite{lin2024decentralized} include a highly simplified point mass vehicle model, an idealized vortical flow model for environmental disturbances, and a perfect perception assumption. 
Realistic wind and wave effects on motion and perception were not considered.
In addition, IQN was implemented with a predefined discrete set of motion commands for vehicle control, and was not able to output arbitrary actions in a continuous action space.
In recent years, Distributional RL algorithms with actor-critic structures \cite{barth2018distributed}, \cite{duan2021distributional}, \cite{singh2022sample} have been developed to work over continuous action domains.
Adopting this idea, we develop a novel ASV decision making and control policy based on IQN employed within an actor-critic framework, which we denote AC-IQN, for continuous control in scenarios involving congested multi-vehicle encounters.

The Convention on the International Regulations for Preventing Collisions at Sea (COLREGs) \cite{international1972convention} specify rules that apply to two-vessel encounter situations, but it faces challenges such as creating conflicting actions in multi-vessel cases \cite{jeong2022motion} and limited maneuvering space near static obstacles like buoys \cite{lin2024decentralized}. 
In addition, the perceived states of nearby objects may not be precise, and can cause false judgements concerning COLREGs. 
To address these issues, we design a reward function that encourages COLREGs compliant behaviors, but also does not penalize other collision avoidance behaviors that are beneficial to navigation safety and efficiency.
The software implementation of our approach is available at {\tt\protect\url{https://github.com/RobustFieldAutonomyLab/Distributional_RL_Decision_and_Control}}.
The main contributions of this letter are summarized as follows:
\begin{itemize}
    \item Proposal of an ASV decision making and control solution based on AC-IQN for navigating in congested multi-vehicle environments, under the influence of wind and wave disturbances on ASV motion and perception.
    \item Design of a novel reward function that trains policies capable of both following and deviating from COLREGs as needed for safe and efficient ASV behavior.
    \item Extensive evaluations in high-fidelity Gazebo simulations show superior navigation safety and efficiency performance over state-of-the-art Distributional RL, non-Distributional RL and classical baselines.
\end{itemize}

The rest of this letter is organized as follows.
A review of relevant literature is given in Section \ref{sec:related works}.
The environment dynamic model of the Gazebo simulator and the vehicle control settings we use are introduced in Section \ref{sec:problem statement}.
Section \ref{sec:perception} introduces the perception module of the proposed system. 
Section \ref{sec:RL approach} describes decision and control with AC-IQN and other RL baselines.
Section \ref{sec:experiments} shows experimental results.
Section \ref{sec:conclusion} concludes the letter and discusses future work.


\vspace{-1.5mm}
\section{Related Works}
\label{sec:related works}

Multi-vessel collision avoidance problems have been widely studied with the consideration of COLREGs.
Naeem et al. \cite{naeem2016reactive} and Chiang et al. \cite{chiang2018colreg} developed COLREGs-compliant algorithms based on Artificial Potential Field (APF) and rapidly-exploring random tree (RRT) respectively by generating virtual obstacles around other vessels to prevent COLREGs-violating actions; Cho et al. \cite{cho2020efficient} developed a rule-based system that specifies the roles of vessels in multi-vessel encounters and utilizes the probabilistic velocity obstacle method for collision avoidance.
However, system performance of these works is only illustrated in highly simplified simulations without modeling of environmental disturbances and perceptual error.
Kuwata et al. \cite{kuwata2013safe} proposed an approach based on Velocity Obstacles (VO) and demonstrated its performance in field tests involving an embodied ASV and manually-driven vessels in an environment with dimensions of about a thousand meters, but the performance in congested scenarios with 
multiple ASVs is not studied.
Hagen et al. \cite{hagen2018mpc} presented a Model Predictive Control (MPC) strategy that chooses actions which minimize the cost of COLREGs-compliant collision avoidance, showing field experiment results involving scenarios with two-vehicle encounters only.

Some works have targeted situations where it is difficult to guarantee navigation safety when following COLREGs, proposing alternative strategies.
Cho et al. \cite{cho2021intent} and Jia et al. \cite{jia2024bayesian} proposed intent inference mechanisms on other encountered vehicles to decides if evasive actions are needed.  
Jeong et al. \cite{jeong2022motion} focused on the multi-ASV collision avoidance problem in highly congested scenarios, and proposed to use multi-objective optimization for action selection based motion attributes of other ASVs.
These works provide insight into the problem where the ego vehicle needs to exhibit different motion patterns based on the scenarios encountered, under the ideal assumption of no wind and wave disturbances and no perceptual error.
Additionally, environments containing static obstacles are not considered in these works.

In recent years, Deep Reinforcement Learning (DRL), given its ability to adapt to complex dynamic scenarios, has been used to develop collision avoidance policies for ASVs assuming no external disturbances to vehicles.
Zhao et al. \cite{zhao2019colregs} presented a DRL policy-gradient agent for ASV path following and COLREGs-compliant collision avoidance task.
Meyer et al. \cite{meyer2020colreg} trained a proximal policy optimization (PPO) based controller that can navigate among other path-following vessels and static obstacles.
Li et al. \cite{li2021path} used APF to improved the performance of a DQN agent in ASV control problems by enabling continuous actions and providing richer reward signals.  
Heiberg et al. \cite{heiberg2022risk} incorporated collision risk indices (CRIs) into the reward function to guide the learning of a COLREGs-compliant RL agent.
Wei et al. \cite{wei2022colregs} proposed a multi-agent reinforcement learning approach that promotes collaborative collision avoidance behaviors among vessels.

\vspace{-1.5mm}
\section{Problem Statement}
\label{sec:problem statement}

In this work, we focus on the autonomous navigation problem of ASVs in relatively dense and congested maritime environments. 
Each ASV is required to navigate to a specified goal location given only measurements from onboard LiDAR and odometry sensors, while avoiding collisions with other surrounding ASVs and buoys.     

We use Virtual RobotX (VRX) \cite{VRX} as the evaluation platform, which is a realistic Gazebo based marine simulation environment developed by Open Robotics \cite{Open_robotics} and used in the Maritime RobotX Competition \cite{RobotX_competition}.
VRX adopts a dynamic model from \cite{fossen2011handbook}, which is shown as follows:
\begin{equation}
\begin{aligned}
    \label{eq:VRX dynamic model}
    &\textbf{\textit{M}}_{RB}\dot{\bm{\nu}} + \textbf{\textit{C}}_{RB}(\bm{\nu})\bm{\nu} + \textbf{\textit{M}}_A \dot{\bm{\nu}}_r + \textbf{\textit{C}}_A(\bm{\nu}_r)\bm{\nu}_r + \\
    &\textbf{\textit{D}}(\bm{\nu}_r)\bm{\nu}_r + \bm{g}(\bm{\eta}) 
    = \bm{\tau} + \bm{\tau}_{\text{wind}} + \bm{\tau}_{\text{wave}}.
\end{aligned}
\end{equation}
In the formula, $\bm{\eta}$ and $\bm{\nu}$ are the 6-dimensional position and velocity vectors. $\bm{\nu}_r$ is the vessel velocity with respect to the fluid. $\bm{\tau}$, $\bm{\tau}_{\text{wind}}$, and $\bm{\tau}_{\text{wave}}$ are the forces and moments from propulsion, wind and waves.

VRX \cite{VRX} models the perturbations of wind and waves on in-water objects, and perception error derives from the need to continually segment and track both static and moving obstacles from raw LiDAR data. 
We use the default settings in VRX \cite{VRX}, where zero-mean Gaussian noise with 0.01 standard deviation is applied to the LiDAR point cloud, and no noise is added to the odometry measurements.

We use the WAM-V ASV model provided by VRX as the deployment vessel, the motion of which is controlled by thrusts from left and right propellers.
We fix the angle of each propeller to zero such that the thrust always aligns with the hull, and the turning motion is achieved by the difference between thrusts.
For each propeller, the maximum forward thrust is set to $1000.0\ \textbf{N}$, and the maximum backward thrust is set to $500.0\ \textbf{N}$. 
We denote the thrust range as $[-500.0,1000.0]\ \textbf{N}$.
With the given thrust range, the vehicle achieves a maximum forward speed of about 3.3 m/s and a maximum backward speed of about 2.3 m/s.

\vspace{-1.5mm}
\section{Perception Processing}
\label{sec:perception}
Given LiDAR and odometry measurements, we obtain the state information of the ego vehicle and objects in its immediate surroundings as shown in the following equations:
\vspace{-1mm}
\begin{equation}
    \vspace{-1.5mm}
    \label{eq:self state}
    \mathbf{s}_{ego} = [p_{x}^{goal},p_{y}^{goal},v_x,v_y,w,T_{left},T_{right}]
\end{equation}
\begin{equation}
    \vspace{-1.5mm}
    \label{eq:object state}
    \mathbf{s}_{object} = [\mathbf{o}_1,\dots,\mathbf{o}_n],\ \mathbf{o}_i=[p^i_x,p^i_y,v^i_x,v^i_y,r_i].
\end{equation}
In the ego state $\mathbf{s}_{ego}$, $p_{x}^{goal}$ and $p_{y}^{goal}$ are the $x$ and $y$ coordinates of the given goal location, $v_x$ and $v_y$ are the $x$ and $y$ components of the linear velocity, $w$ is the yaw component of the angular velocity, and $T_{left}$, $T_{right}$ are thrusts of the left and right propellers. 
To simplify the perception, we consider each perceived object to be a cylindrical shape.  
The state of the $i$-th perceived object consists of its position, $[p^i_x,p^i_y]$, velocity, $[v^i_x,v^i_y]$, and radius $r_i$.
Both ego state and object states are expressed in the frame of the ego vehicle. 

The algorithm we use to extract object information from the LiDAR point cloud is based on \cite{bogoslavskyi2016fast} and described in Algorithm \ref{Alg:LiDAR segmentation}, where $R(\cdot)$ computes the distance between a LiDAR point and the LiDAR center, $M(\cdot)$ and $m(\cdot)$ are the max and min operators respectively, and $\theta$ is a given threshold to decide whether a point belongs to a cluster.
An example of point cloud segmentation is shown in Figure \ref{fig:system}, where different clusters are shown in different colors.
Given that velocity estimates of distant clusters are less reliable due to the influence of environmental disturbances, we discard LiDAR reflections beyond 20 meters.

After segmentation, each cluster is regarded as an object, and the centroid and radius of the cluster are used as the position and radius of the object.
To estimate the velocity of the object, we project the clusters from the vehicle frame at the last time step to the current frame using a relative transformation computed with odometry measurements.
For simplicity, we assume that a perceived object with estimated speed smaller than 0.5 m/s is not a vehicle and we do not include it in the COLREGs checking process mentioned in Section \ref{sec:RL approach}. 
The resulting ego state and object states will be used in the decision making and control module.
\begin{algorithm}
\hspace{-2mm}{\bf Input}: LiDAR point clouds $P$, {\bf Output}: Clusters set $C$ \\ 
    $i=0$, $C[i].centroid=0.0$, $C[i].raidus=0.0$ \\
    {\bf for} each point $j$ in $P$ \\
    \hspace{4mm}{\bf if} $j$ belongs to an existing cluster {\bf then continue}  \\
    \hspace{4mm}Initialize a new queue $q\leftarrow\emptyset$, $q.push(j)$, $i = i + 1$\\
    \hspace{4mm}{\bf while} $q$ is not empty \\
    \hspace{8mm}$k=q.top()$, $C[i].add(k)$, $q.pop()$ \\
    \hspace{8mm}Update $C[i].centroid$ and $C[i].raidus$ with $k$ \\
    \hspace{8mm}{\bf for} each point $p$ that is a neighbor of $k$ \\
    \hspace{12mm}$d_1 = M(R(p),R(k)), d_2 = m(R(p),R(k))$ \\
    \hspace{12mm}{\bf if} $\arctan\frac{d_2 \sin{\alpha}}{d_1 - d_2 \cos{\alpha}} > \theta$ {\bf then} $q.push(k)$
\caption{LiDAR point cloud segmentation}
\label{Alg:LiDAR segmentation}
\end{algorithm}
\setlength{\textfloatsep}{5pt}

\section{Decision Making and Control with RL}
\label{sec:RL approach}
\subsection{Problem Formulation}
A Markov Decision Process $(\mathcal{S},\mathcal{A},P,R,\gamma)$ is used to describe the ASV navigation problem, where $\mathcal{S}$ and $\mathcal{A}$ are the state space and action space of agents interacting with the environment.
$P(\cdot|s,a)$ is the state transition function describing the evolution of environment state, which reflects the dynamics in Equation \eqref{eq:VRX dynamic model}.
The reward function $R(s,a)$ generates scalar value signals that indicate the preference towards state-action pairs.
At each time step $t$, based on the observation of the current state $s_t$, each agent selects an action $a_t$ according to the policy $\pi$. 
Then the environment transitions to the next state $s_{t+1}$ according to $P(\cdot|s_t,a_t)$ and each agent receives a reward $r_{t+1}=R(s_{t+1},a_{t+1})$.  
The action value function $Q^\pi(s, a)$ is defined as the expected cumulative reward of taking action $a$ at state $s$ and following the policy $\pi$ thereafter, where the discount factor $\gamma\in[0,1)$ reflects the importance of future rewards.
\vspace{-1mm}
\begin{equation}
  \vspace{-1mm}
  \label{eq:expected return}
  Q^\pi(s, a) = \mathbb{E}_\pi[\sum\nolimits_{k=0}^\infty \gamma^k r_{t+k+1}| s_t=s, a_t=a]
\end{equation}

The objective is to obtain an optimal policy $\pi_*$ that maximizes $Q^\pi(s,a)$ for all state-action pairs, and the resulting optimal action value function $Q^{\pi_*}(s, a)$ satisfies the Bellman optimality equation $Q^{\pi_*}(s, a) = \mathbb{E}[r_{t+1}+\gamma\max_{a'}Q^{\pi_*}(s', a')]$.

\vspace{-2mm}
\subsection{Action Commands}
\label{subsec:action commands}

The action commands in each control time step are the variations in the propellers' thrusts.
For methods that output discrete actions, the action command of each propeller is chosen from a given action set.
We observed that increasing the number of actions in the discretized action space causes difficulties in learning stable control policies, and chose the action set $\mathbf{A}=\{-1000.0, -500.0, 0.0, 500.0, 1000.0\}\ \textbf{N} / \text{s}$.
For methods that operate in continuous action space, each action $a\in[-1000.0,1000.0]\ \textbf{N} / \text{s}$.
When applying actions, thrusts are clipped within the thrust range noted in Sec. \ref{sec:problem statement}.

\vspace{-2mm}
\subsection{Deep Reinforcement Learning}
\label{subsection: deep RL}
DQN \cite{mnih2015human} uses a deep neural network model to approximate the action value function $Q(s,a)$, and train it by optimizing the loss based on Temporal Difference (TD) error.
\vspace{-2mm}
\begin{equation}
    \vspace{-1mm}
    \label{eq:DQN loss}
    \mathcal{L}_{\text{DQN}} = \mathbb{E}[(r+\gamma\max_{a'}Q(s',a';\theta^-)-Q(s,a;\theta))^2]
\end{equation}
Rainbow \cite{hessel2018rainbow} outperforms DQN in the Atari 2600 benchmark by incorporating techniques that suppress overestimation and improve stability and efficiency in learning value functions.
For continuous control problems, the action space needs to be discretized for DQN and Rainbow to be applied.

DDPG \cite{lillicrap2015continuous} is an actor-critic method that is applicable to continuous action space.
It maintains a critic model $Q(s,a|\theta^Q)$ that can be learned with a loss function similar to Equation \eqref{eq:DQN loss}, and an actor model $\mu(s|\theta^\mu)$, the parameters of which are updated using the following policy gradient.
\vspace{-1mm}
\begin{equation}
    \vspace{-1mm}
    \label{eq:DDPG policy gradient}
    \nabla_{\theta^\mu}J\approx\mathbb{E}[\nabla_aQ(s,a|\theta^Q)|_{a=\mu(s)}\nabla_{\theta^\mu}\mu(s|\theta^\mu)]
\end{equation}
SAC \cite{haarnoja2018soft} further improves exploration with an objective function \eqref{eq:SAC objective} that involves the entropy of the learned policy, and uses a stochastic actor with enhanced robustness.
\vspace{-1mm}
\begin{equation}
    \label{eq:SAC objective}
    J=\sum\nolimits_{t=0}^T\mathbb{E}[r(s_t,a_t)+\alpha\mathcal{H}(\pi(\cdot|s_t))]
\end{equation}

\vspace{-1.5mm}
\subsection{Distributional Reinforcement Learning}
\label{subsection: Distributional RL}

Instead of the expected return $Q^\pi(s,a)$, Distributional RL algorithms \cite{bellemare2017distributional} learn the return distribution $Z^\pi(s,a)$, where $Q^\pi(s,a) = \mathbb{E}[Z^\pi(s,a)]$, and the distributional Bellman equation, $Z^\pi(s,a) \overset{D}{=} R(s,a) + \gamma Z^\pi(s',a') $, is considered.
When the dynamics of the operating environment are highly uncertain, 
the randomness of the collected reward samples adversely affect the accuracy of the learned action value.
Distributional RL methods can mitigate this problem by modeling and learning the distribution of cumulative rewards.

Implicit Quantile Networks (IQN) \cite{dabney2018implicit} express the return distribution with a quantile function $Z_\tau := F_Z^{-1}(\tau)$, where $\tau\sim U([0,1])$, and represents the policy as follows.
\vspace{-2mm}
\begin{equation}
    \vspace{-1mm}
    \label{eq:approximate pi}
    \pi(s) = \text{argmax}_a \frac{1}{K}\sum\nolimits_{k=1}^K Z_{\tau_k}(s,a),\tau_k\sim U([0,1])
\end{equation}
Parameters of the IQN policy model can be learned by optimizing the loss defined in Equation \eqref{eq:IQN loss}.
The outputs of the quantile function, $Z_{\tau_k}$, are referred to as action quantile values.
IQN also requires a discretized action space.
\vspace{-2mm}
\begin{equation}
    \vspace{-1mm}
    \label{eq:sampled TD}
    \delta^{\tau_i,\tau'_j} = r + \gamma Z_{\tau'}(s',\pi(s')) - Z_{\tau}(s,a)
\end{equation}
\begin{equation}
    \vspace{-1mm}
    \label{eq:quantile Huber}
    \begin{array}{c}
        \rho_\tau^\kappa(u) = |\tau-\mathbf{1}_{\{u<0\}}|(\mathcal{L}_\kappa(u)/\kappa), \\[5pt]

        \text{where }\mathcal{L}_\kappa(u) = \left\{
            \begin{array}{lr}
                 \frac{1}{2}u^2, &\text{if}\ |u|\leq\kappa \\
                 \kappa(|u|-\frac{1}{2}\kappa), & \text{otherwise} 
            \end{array}
        \right.
    \end{array}
\end{equation}
\begin{equation}
    \label{eq:IQN loss}
    \mathcal{L}_{\text{IQN}} = \frac{1}{N'}\sum\nolimits_{i=1}^N \sum\nolimits_{j=1}^{N'} \rho_{\tau_i}^\kappa(\delta^{\tau_i,\tau'_j})
\end{equation}

\vspace{-1.5mm}
\subsection{Actor Critic Implicit Quantile Networks}

In this work, we develop a novel decision making and control policy for ASV navigation based on IQN within an actor-critic framework, which we denote AC-IQN.
Compared to policies based on traditional RL in Sec. \ref{subsection: deep RL}, the proposed policy uses Distributional RL, which can be more robust to the marine operating environment and its uncertainties from motion disturbances, perception errors and unknown intents of other vehicles.
The proposed policy also exhibits superior maneuverability over the policy based on IQN in Sec. \ref{subsection: Distributional RL}, which is crucial for deployment in congested scenarios.

\begin{algorithm}
    \hspace{-2mm}{\bf Input}: Critic $Z_{\theta}$, Actor $\pi_{\phi}$, replay buffer $\mathcal{M}$ \\
    Sample a minibatch $\{(s,a,r,s')_i\}_{i=1}^{M}$ from $\mathcal{M}$ \\
    Compute gradient of the critic loss $\nabla_{\theta} L$, where
    \begin{align*}
        L = &\frac{1}{M}\sum_{i=1}^M[\frac{1}{N'}\sum_{i=1}^N \sum_{j=1}^{N'} \rho_{\tau_i}^\kappa(r + \gamma Z_{\theta'}^{\tau_j}(s',\pi_{\phi'}(s')) \\ &- Z_{\theta}^{\tau_i}(s,a))]
    \end{align*}
    Compute the policy gradient $\nabla_{\phi} J$, where
    \begin{align*}
        \nabla_{\phi} J = \mathbb{E}[\frac{1}{N}\sum_{i=1}^N \nabla_{a} Z_{\theta}^{\tau_i}(s,a)|_{a=\pi_{\phi}(s)} \nabla_{\phi} \pi_{\phi}(s)]
    \end{align*}
    Update critic and actor network parameters
    \begin{align*}
        \theta\leftarrow \theta + \alpha_{\theta}\cdot \nabla_{\theta} L,\ \phi\leftarrow \phi + \alpha_{\phi} \cdot \nabla_{\phi} J
    \end{align*}
    Update target network parameters every $k$ iterations
    \begin{align*}
        \theta'\leftarrow \beta\theta + (1-\beta)\theta',\ \phi'\leftarrow \beta\phi + (1-\beta)\phi'
    \end{align*}
\caption{AC-IQN model update}
\label{Alg:AC-IQN}
\end{algorithm}
\begin{figure}
    \vspace{-4mm}
    \centering
    \includegraphics[width=\linewidth]{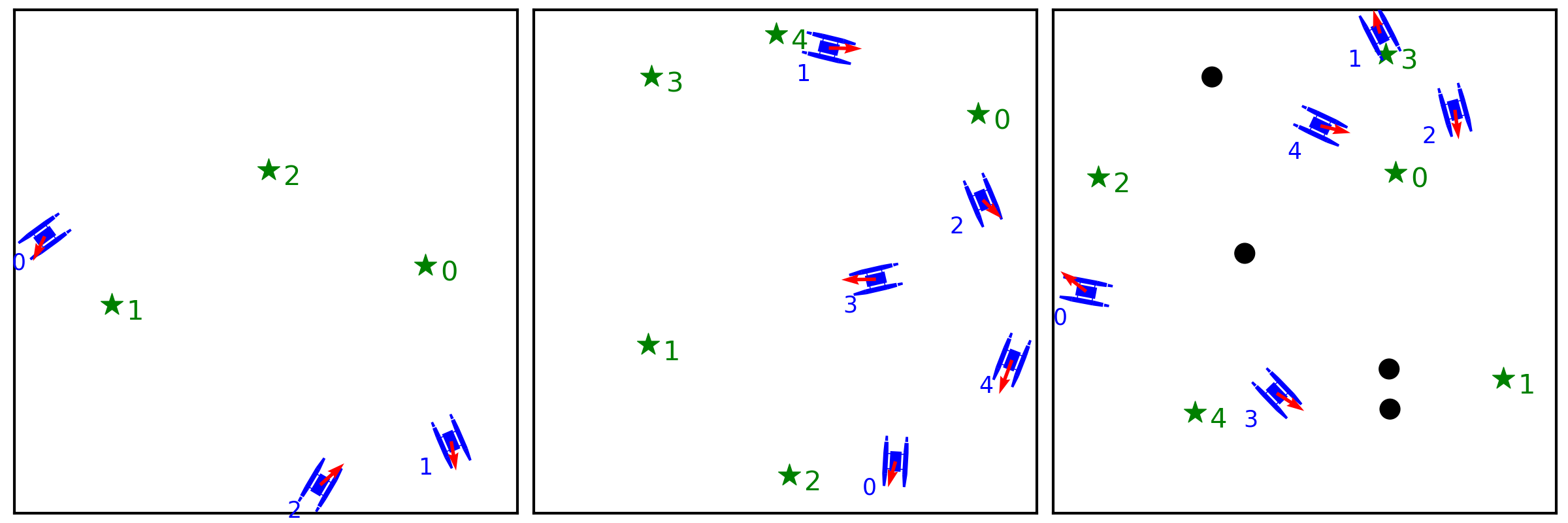}
    \vspace{-5mm}
    \caption{{\bf Example training scenarios.} The velocity of each vehicle is indicated by the red arrow, goal positions are plotted as green stars, and buoys are shown as black circles.}
    \label{fig:training scenarios}
    \vspace{-3mm}
\end{figure}
\begin{table}[h!]
    \centering
    \caption{\textbf{Curriculum training process.}}
    \vspace{-1mm}
    \begin{tabular}{ccccccc}
         \toprule
         Timesteps (million) & 1st & 2nd & 3rd & 4st & 5st & 6st  \\ \midrule
        Number of robots & 3 & 4 & 5 & 5 & 5 & 5 \\ 
        Number of buoys & 0 & 0 & 0 & 2 & 3 & 4 \\ 
        Min distance between & \multirow{2}{*}{30.0} & \multirow{2}{*}{35.0} & \multirow{2}{*}{40.0} & \multirow{2}{*}{40.0} & \multirow{2}{*}{40.0} & \multirow{2}{*}{40.0} \\
        start and goal & & & & & & \\ \bottomrule
    \end{tabular}
    \label{tab:schedule}
\end{table}
The learning algorithm of the proposed decision making and control policy based on AC-IQN is shown in Algorithm \ref{Alg:AC-IQN}.
The Critic $Z_\theta$ is similar to the original IQN policy model, which outputs action quantile values given quantile samples and a state-action pair. 
When computing the Critic loss gradient $\nabla_{\theta}L$, the policy output of the Actor $\pi_{\phi'}$, and quantile samples $\{\tau_i|i=1,\dots,N\}$ and $\{\tau_j|j=1,\dots,N'\}$, drawn from the uniform distribution $U([0,1])$, are used.
The loss function $\rho^{\kappa}_{\tau}$ is defined in Equation \eqref{eq:quantile Huber}.

Policy gradient $\nabla_{\phi}J$ is used to update the Actor $\pi_{\phi}$, which can be computed according to the chain rule as shown in Algorithm \ref{Alg:AC-IQN}. 
The expectation over state is approximated with those from the sampled minibatch.

\begin{figure*}
    \centering
    \includegraphics[width=\linewidth]{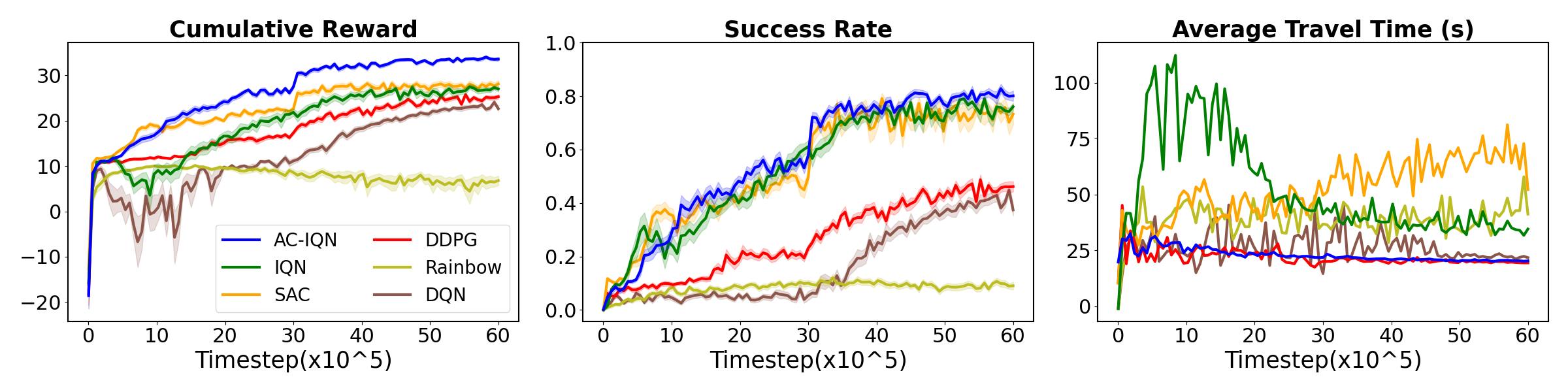}
    \vspace{-8mm}
    \caption{{\bf Learning performance.} Each curve and its band width in the above cumulative reward and success rate plots reflect the values of the mean and standard error. To compute the average travel time, we only include data from robots that successfully reach their goals.}
    \label{fig:learning curves}
    \vspace{-6mm}
\end{figure*}
\begin{figure}
    \centering
    \includegraphics[width=\linewidth]{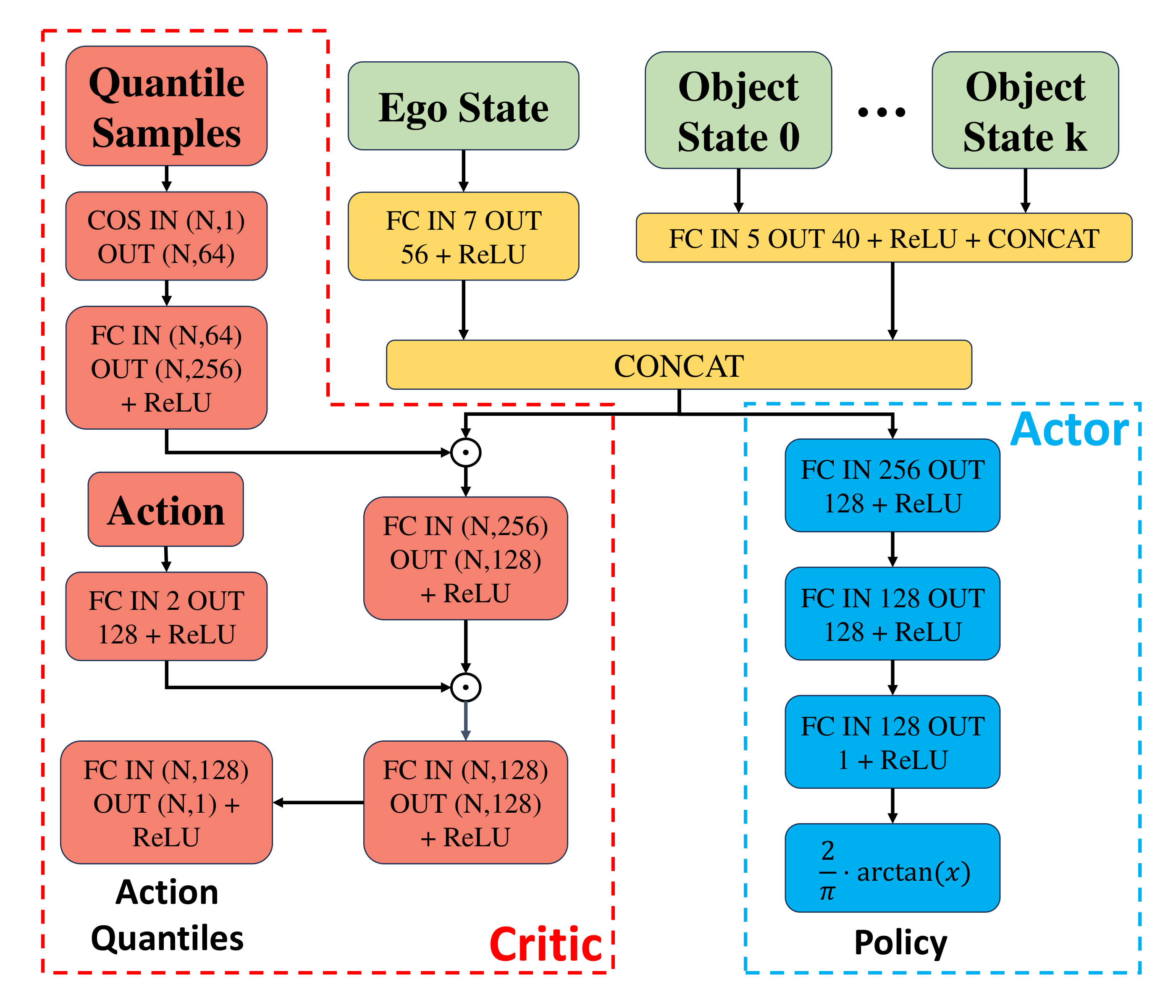}
    \vspace{-7mm}
    \caption{{\bf AC-IQN network architecture.} FC, ReLU, COS, $\odot$ and CONCAT stand for Fully Connected Layer, Rectified Linear Unit, Cosine Embedding Layer, element-wise product and concatenation of tensors. 
    The numbers after IN and OUT are the input and output dimension of a layer.
    }
    \label{fig:network}
\end{figure}

\vspace{-1.5mm}
\subsection{Training RL agents}

We trained our AC-IQN agent, as well as IQN, SAC, DDPG, Rainbow and DQN agents which serve as comparative baselines, and used them in experiments described in Section \ref{sec:experiments}.
The SAC and Rainbow agents are based on the implementations of \cite{weng2022tianshou} and \cite{rainbow_github}.  
Due to the high computational expense of Gazebo's realistic environment simulation, RL agents were trained in a simplified 2D environment.
As shown in Equation \eqref{eq:training dynamic model}, a simplified three Degree-of-Freedom (DoF) dynamic model described in \cite{fossen2011handbook} is used, which only includes surge, sway, and yaw DoFs. 
\vspace{-1mm}
\begin{equation}
    \label{eq:training dynamic model}
    \textbf{\textit{M}}_{RB}\dot{\bm{\nu}} + \textbf{\textit{C}}_{RB}(\bm{\nu})\bm{\nu} + \textbf{\textit{M}}_A \dot{\bm{\nu}}_r + \textbf{\textit{N}}(\bm{\nu}_r)\bm{\nu}_r
    = \bm{\tau} + \bm{\tau}_{\text{wind}} + \bm{\tau}_{\text{wave}}
\end{equation}

To maintain efficient training processes, the perceptual information defined in Eqs. \eqref{eq:self state} and \eqref{eq:object state} was directly given, and $\bm{\tau}_{\text{wind}}$ and $\bm{\tau}_{\text{wave}}$ were assumed to be zero.
We introduced noisy perception, defined in Eqs. \eqref{eq:pos noise} - \eqref{eq:r noise}, to increase the robustness of trained RL agents to motion disturbances and perception errors.
\vspace{-2mm}
\begin{equation}
    \vspace{-2mm}
    \label{eq:pos noise}
    P_{o} = P + w_p,\ w_p\sim\mathcal{N}(0,\Sigma_p)
\end{equation}
\begin{equation}
    \vspace{-2mm}
    \label{eq:vel noise}
    V_{o} = V + w_v,\ w_v\sim\mathcal{N}(0,\Sigma_v)
\end{equation}
\begin{equation}
    \vspace{-1mm}
    \label{eq:r noise}
    R_{o} = R \cdot (r_{mean} + (1-r_{mean})\cdot w_r/\pi),\ w_r\sim\mathcal{V}(0,\kappa) 
\end{equation}
In the above equations, $P$, $V$, and $R$ are the ground truth position, velocity and radius of the perceived objects. Position noise $w_p$ and velocity noise $w_v$ are drawn from zero-mean Gaussian distributions. Since the perceived radius is bounded by the actual enclosing radius of the object, we model the radius noise $w_r$ with a von Mises distribution, which is a close approximation to the wrapped normal distribution and lies within $[-\pi,\pi]$.
As the training schedule in Table \ref{tab:schedule} shows, the complexity of the randomly generated training environment gradually increases as the training proceeds. 
Example training scenarios are visualized in Figure \ref{fig:training scenarios}.
Similar to our prior work \cite{lin2024decentralized}, we only maintain one RL model during training, which is shared with all vehicles for individual decision making and control tasks.
\begin{figure}
    \centering
    \includegraphics[width=\linewidth]{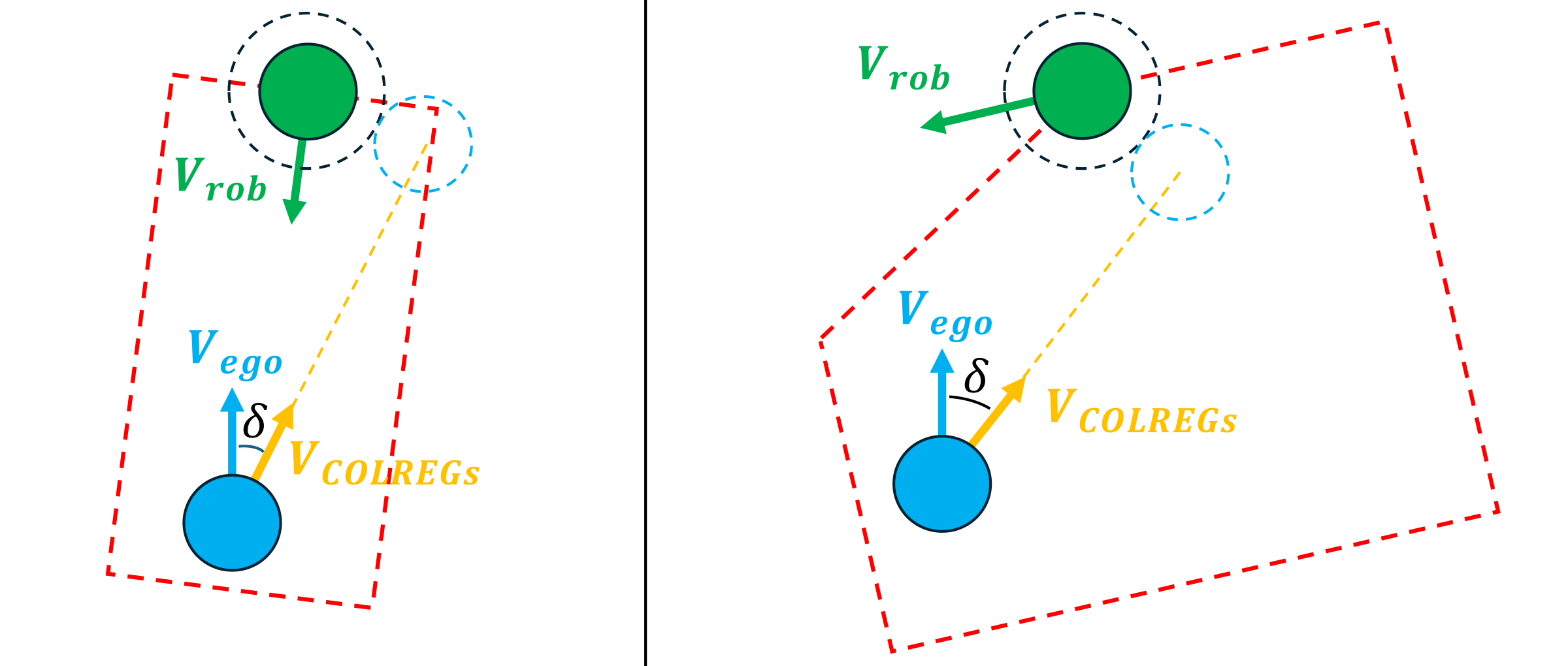}
    \vspace{-5mm}
    \caption{{\bf Head-on and crossing scenarios.} Velocities of the ego vehicle that are consistent with COLREGs requirements are plotted in yellow. 
    The COLREGs compliant velocity of each vehicle is computed by viewing it as the ego vehicle.}
    \label{fig:COLREGs}
\end{figure}
\vspace{-1mm}
\begin{equation}
    \begin{aligned}
    r_t &= r_{\text{step}} + r_{\text{forward}} +  r_{\text{COLREGs}}\cdot I(s_t\in\mathcal{S}_{\text{COLREGs}}) \\
    &+r_{\text{collision}}\cdot I(s_t\in\mathcal{S}_{\text{collision}}) + r_{\text{goal}}\cdot I(s_t\in\mathcal{S}_{\text{goal}}) 
    \end{aligned}
    \label{eq:reward}
\end{equation}

The reward function we use in training is shown in Equation \eqref{eq:reward}, where $I(\cdot)$ is the indicator function, $r_{\text{step}}=-0.1$, $r_{\text{forward}}=||p^{goal}_{t-1}||-||p^{goal}_{t}||$, $r_{\text{collision}}=-5.0$, $r_{\text{goal}}=10.0$.
We considered three cases of two-vehicle encounters where each vehicle shall follow the behavior specified by COLREGs \cite{international1972convention} unless other actions are needed to avoid collisions: (1) Overtaking; (2) Head-on; (3) Crossing.  
For case (1), the vessel at the back shall move out of the way of the vessel being overtaken.
$r_{\text{collision}}$ penalizes actions leading to collisions, and thus promotes behaviors that conform to the rule of case (1).
For case (2), each vessel shall alter course to starboard and pass on the port side of the other.
For case (3), the vessel that has the other on its starboard side shall give way.
To promote behaviors that follow the rules of cases (2) and (3), which are visualized in Figure \ref{fig:COLREGs}, each vehicle, $ego$, will be checked against its closest vehicle, $rob$: 1) 
$ego$ is determined to be in a head-on situation if it is in the head-on zone of $rob$, and the absolute value of the angle from $V_{rob}$ to $V_{ego}$ is greater than $3\pi/4$; 2) $ego$ is the give way vehicle in the crossing situation if it is in the crossing zone of $rob$, and the angle from $V_{rob}$ to $V_{ego}$ is in $[\pi/4,3\pi/4]$.
If $ego$ is in situation 1) or 2), which we denote as $s_t\in\mathcal{S}_{\text{COLREGs}}$, then $r_{\text{COLREGs}} = -0.1\cdot\max(\delta,0.0)$ will be applied to $ego$, where $\delta\in[-\pi,\pi)$ is the angle from $V_{ego}$ to $V_{\text{COLREGs}}$ (clockwise is positive).
$\mathcal{S}_{\text{collision}}$ and $\mathcal{S}_{\text{goal}}$ are situations where $ego$ collides with other objects and reaches the goal respectively.

            
The network structure for the AC-IQN agent is shown in Figure \ref{fig:network}. 
The Cosine Embedding Layer computes a feature $[\cos(\pi\cdot 0\cdot \tau),\dots,\cos(\pi\cdot 63\cdot \tau)]$ for each quantile sample $\tau$.
The outputs of Actor are numbers in $(-1.0,1.0)$, which are then multiplied by $1000.0$ to match the action range.
The structure of the networks used by IQN, DDPG and DQN agents are mostly the same as AC-IQN, except for necessary modifications to match the corresponding input and output.

The learning performances of the AC-IQN, IQN, SAC, DDPG, Rainbow and DQN agents are shown in Figure \ref{fig:learning curves}.
Considering the effect of random seed selection on training, we train 30 models with different random seeds for each agent on an Nvidia RTX 3090 GPU and show the general performance. 
AC-IQN shows a clear advantage in the cumulative reward obtained during the training process relative to other methods, and achieves the highest success rate with a minimal amount of travel time.
IQN and SAC achieve similar levels of safety performance as AC-IQN, but at the cost of longer average travel time in general.
On the contrary, DDPG and DQN can complete navigation tasks efficiently, but have much lower success rates than AC-IQN. 
Rainbow does not succeed in learning safe and efficient policies, which may result from the effects of substantial observation noise and uncertainty in multi-vehicle interactions on the performance of Rainbow components like multi-step return and prioritized replay.

 \vspace{-1mm}
\section{Experiments}
\label{sec:experiments}
Each navigation system evaluated in our experiments requires the integration of the perception module introduced in Sec. \ref{sec:perception} and a decision making \& control agent into the Robot Operating System (ROS 2) framework, and we use all six RL methods described in Sec. \ref{sec:RL approach} in the experiments.
For each RL method, we choose the model with average
training success rate performance of all trained models for use in our experiments.   
In addition to RL agents, we also include two classical methods, Artificial Potential Fields (APF) and Model Predictive Control (MPC), which also consider COLREGs in ASV navigation and are described in the following paragraphs, as baselines. 

\begin{figure}
    \centering
    \includegraphics[width=\linewidth]{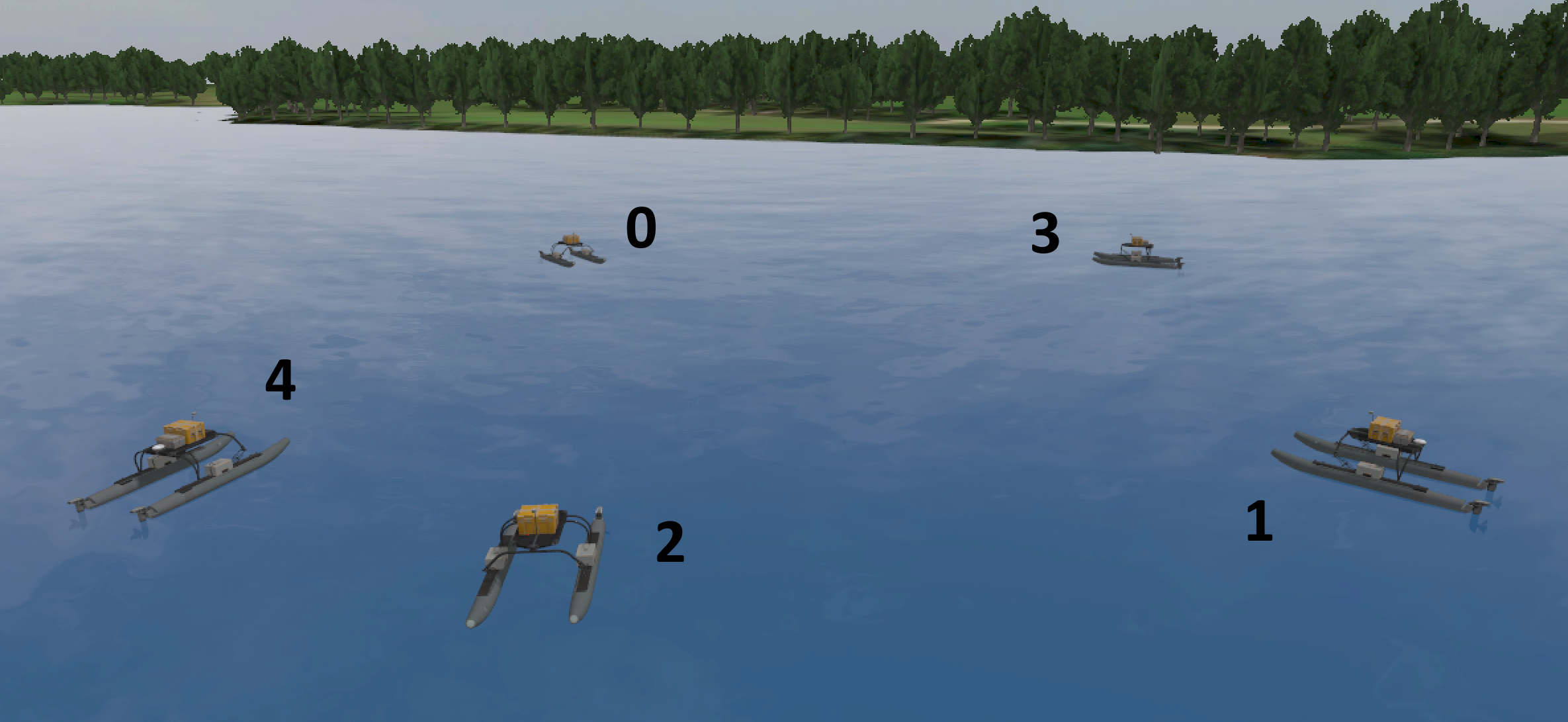}
    \includegraphics[width=\linewidth]{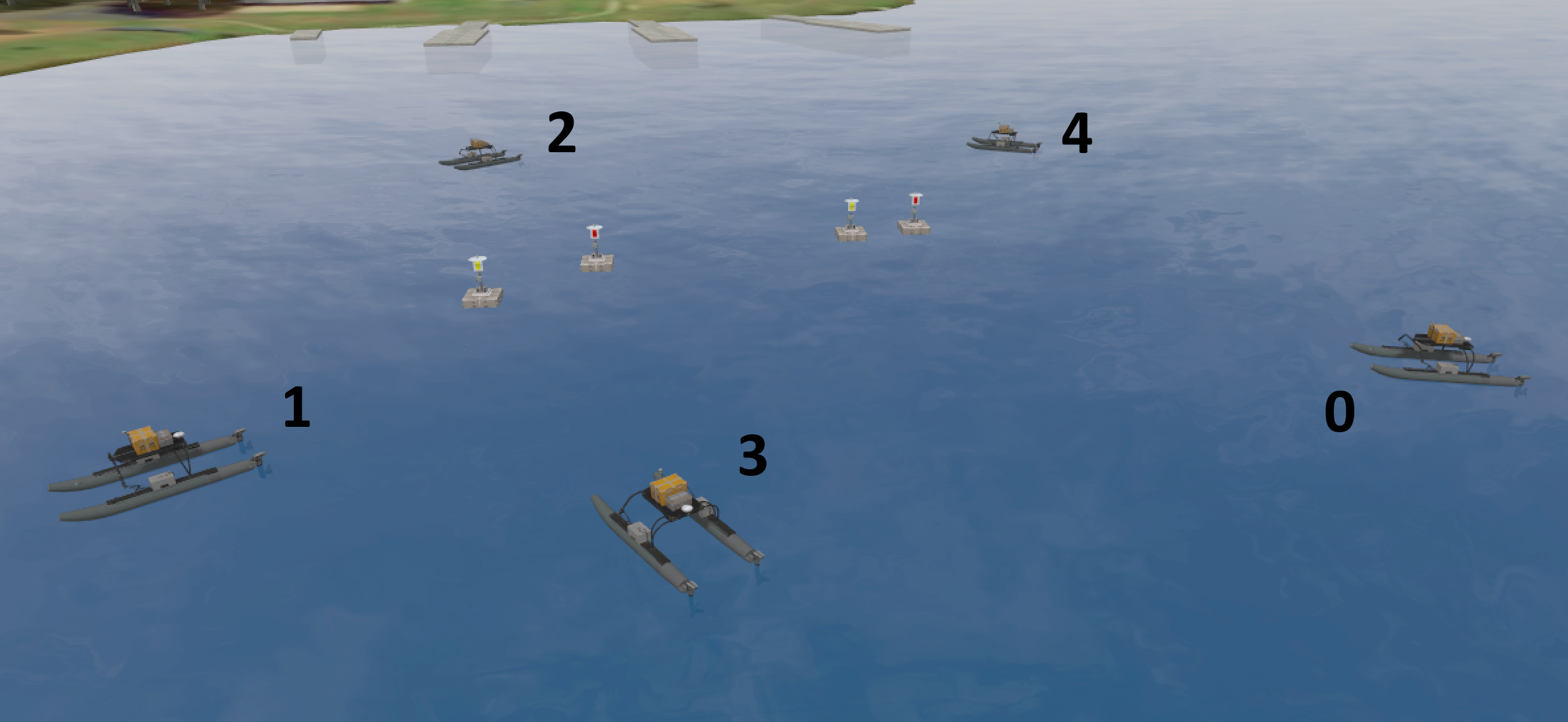}
    \caption{{\bf Example VRX experiment episodes (with and without buoys).} The initial vehicle poses and the buoy positions are shown.}
    \label{fig:vrx exp setup}
\end{figure}

The experiments were run on an AMD Ryzen threadripper 3970X CPU and used the $\texttt{sydney\_regatta}$ environment of the VRX simulator.
We approximate the scenario of Fig. 5 in \cite{VRX} and set the average wind speed to 10 m/s. 
The wave parameters of period and gain ratio, defined in \cite{VRX}, are respectively set to 5.0 sec. and 0.3 to approximate the conditions of the inland water at $\texttt{sydney\_regatta}$.
We performed six sets of experiments with an increasing level of complexity reflected in the number of vehicles and buoys, where each set includes 100 experiments for each navigation system.
Examples of experiment scenarios are visualized in Figure \ref{fig:vrx exp setup}.
For an experiment episode, all vehicles are equipped with the same navigation system, and the episode is considered failed if a collision happens or the travel time of any vehicle exceeds 90 seconds.

The APF agent we implemented is based on \cite{fan2020improved} and \cite{naeem2016reactive}.
COLREGs-compliant behaviors are promoted by generating virtual obstacles to prevent vehicles from entering COLREGs-violating locations.
The repulsive forces $U_{\text{rep}} = 0.5\cdot k_{\text{rep}}\cdot(1/d(X,X_o)-1/d_0)^2\cdot d^2(X,X_g)$ are generated when $d(X,X_o)\leq d_0$.
$X$ and $X_g$ are positions of the robot and goal, and $X_o$ are the positions of perceived objects and virtual obstacles.
The attractive force $U_{\text{att}}(X) = 0.5 \cdot k_{\text{att}} \cdot d^2(X,X_g)$.
We use $k_{\text{att}}=50.0$, $k_{\text{rep}}=500.0$, $d_0=15.0$,
and total force $F = -\nabla U_{\text{att}}(X)-\nabla U_{\text{rep}}(X)$.
To map the force to an action command, we forward simulate all combinations of action commands from the action set $\mathbf{A}$ defined in Sec. \ref{subsec:action commands} for one control time step, and choose the one that minimizes the difference between velocity at the next step and $F$.
\begin{figure*}
    \centering
    \begin{subfigure}{0.3\textwidth}
        \centering
        \includegraphics[width=\linewidth]{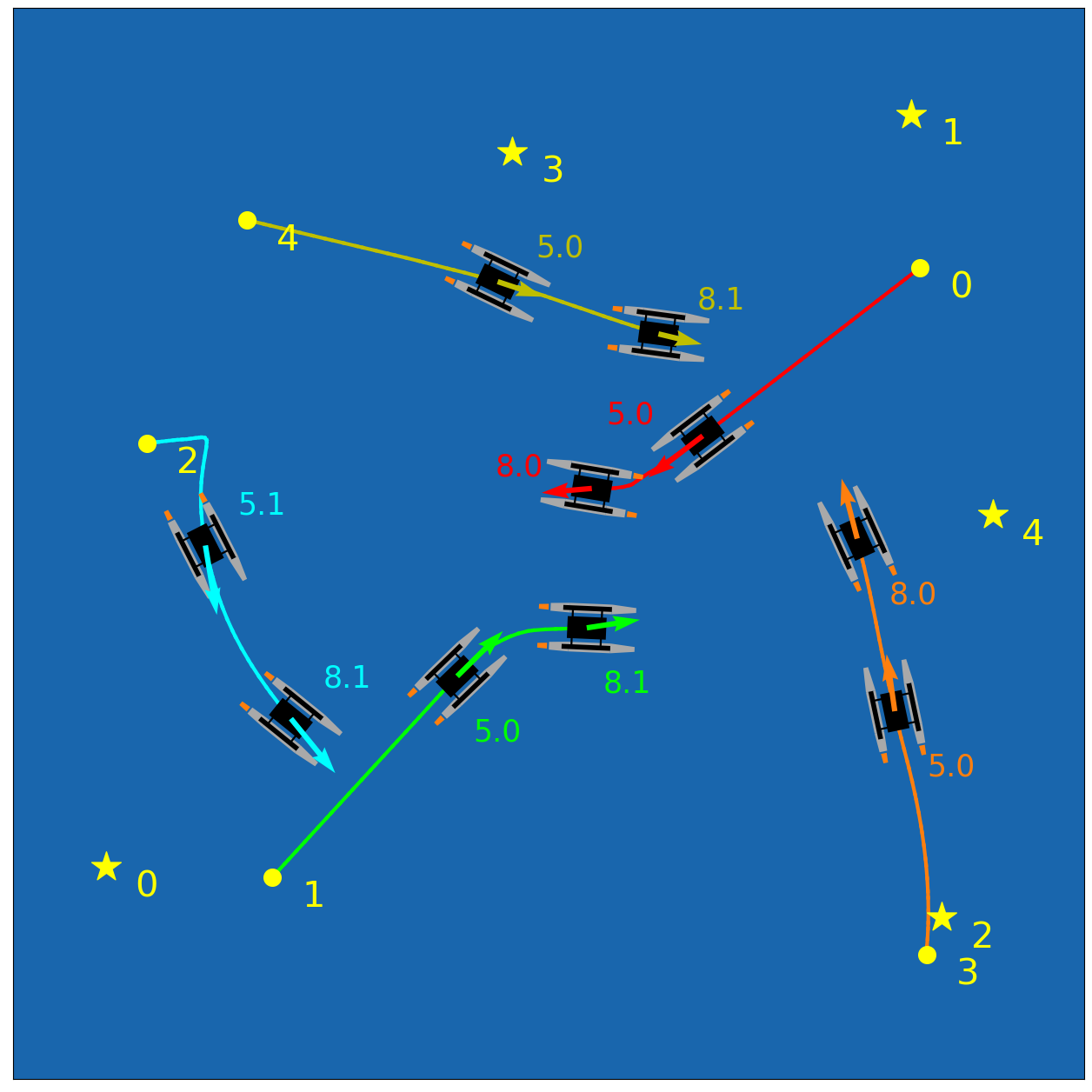}
    \end{subfigure}
    \begin{subfigure}{0.3\textwidth}
        \centering
        \includegraphics[width=\linewidth]{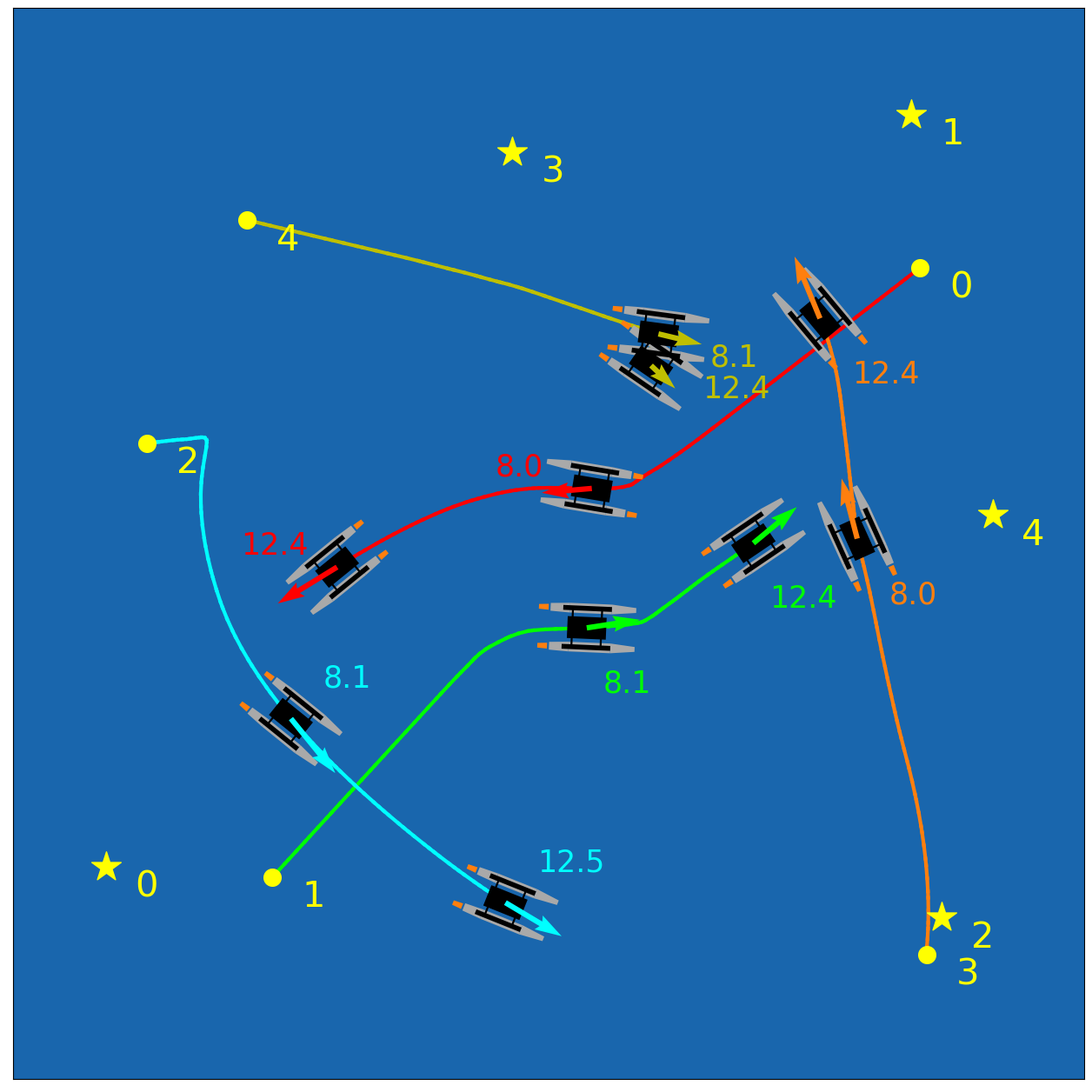}
    \end{subfigure}
    \begin{subfigure}{0.3\textwidth}
        \centering
        \includegraphics[width=\linewidth]{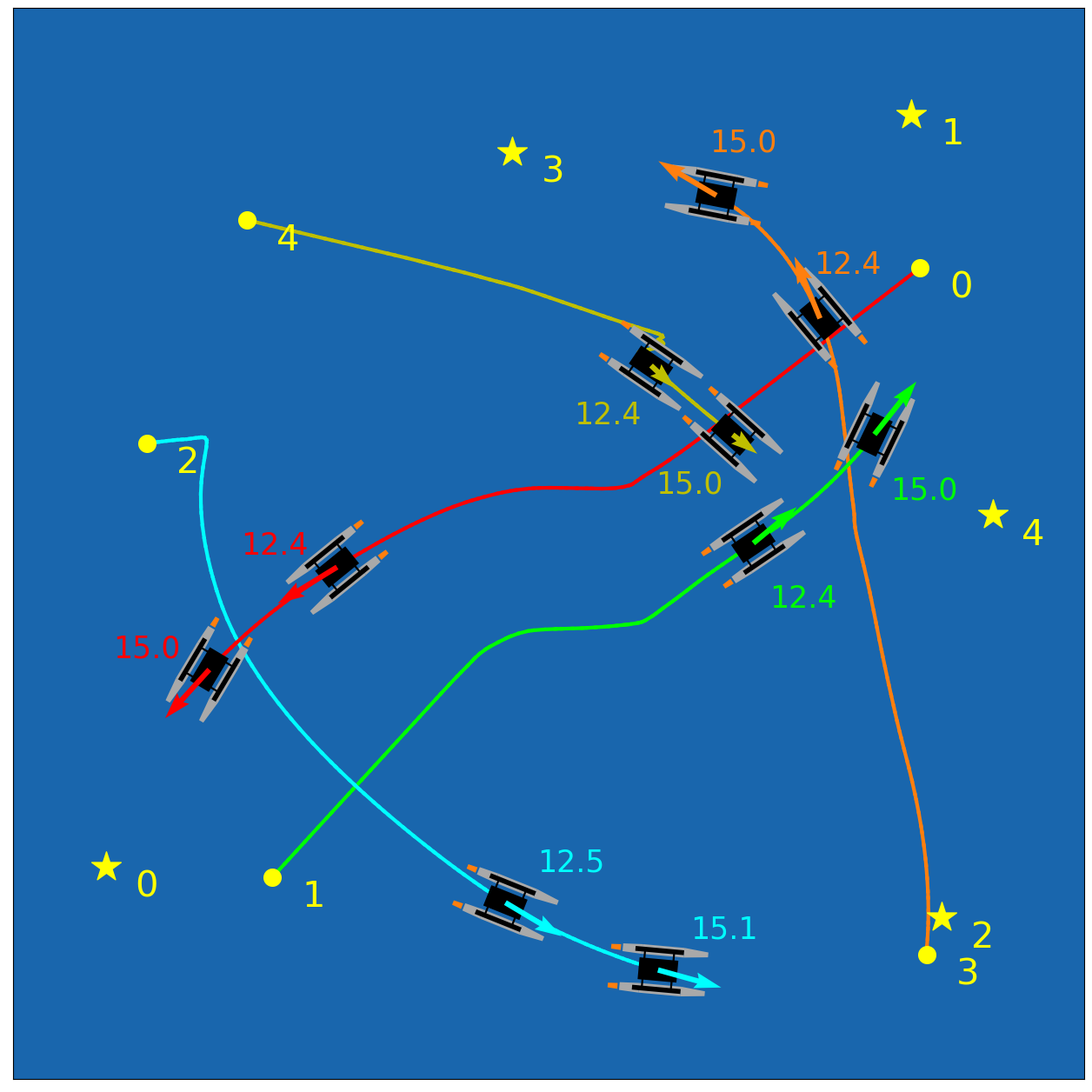}
    \end{subfigure}
    \begin{subfigure}{0.3\textwidth}
        \centering
        \includegraphics[width=\linewidth]{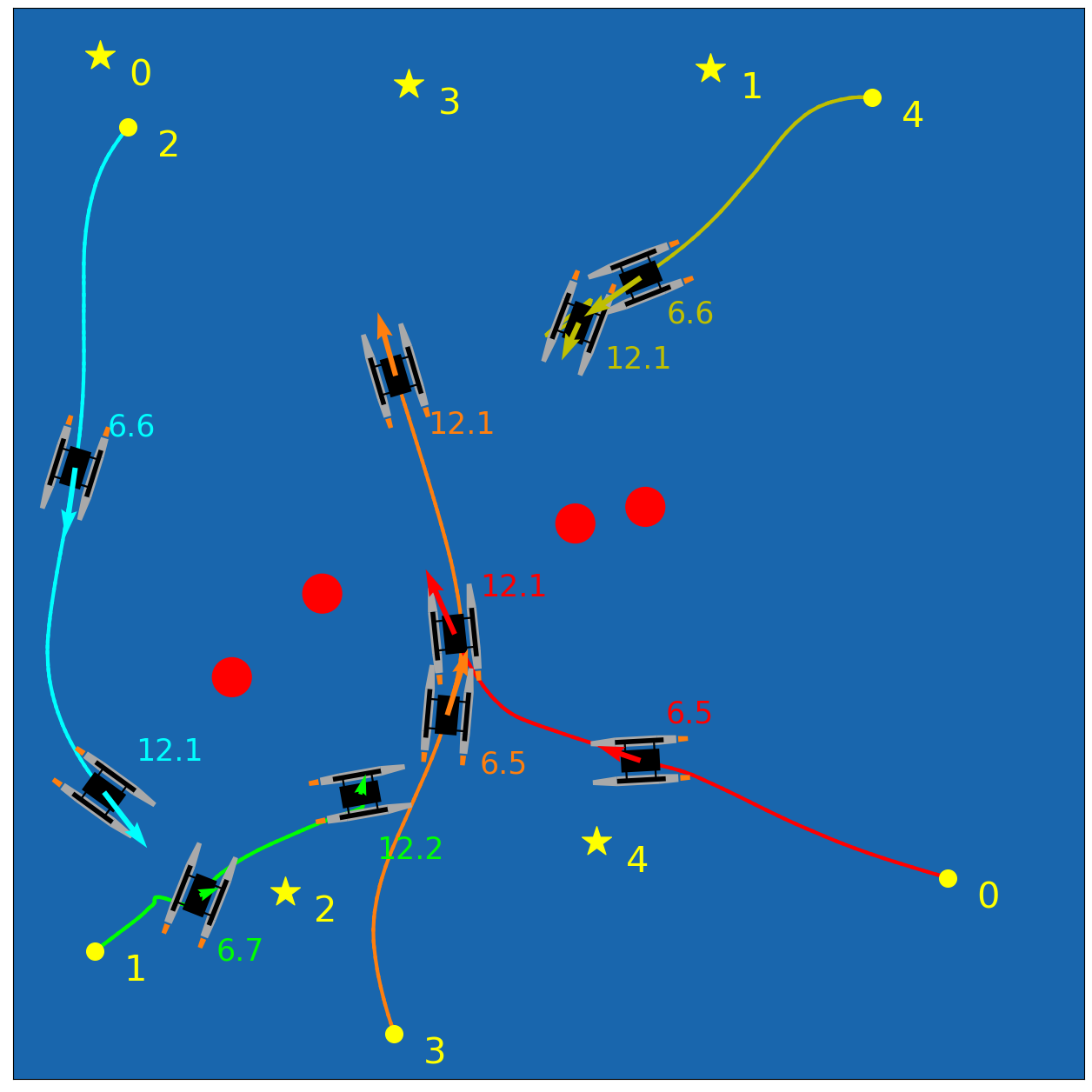}
    \end{subfigure}
    \begin{subfigure}{0.3\textwidth}
        \centering
        \includegraphics[width=\linewidth]{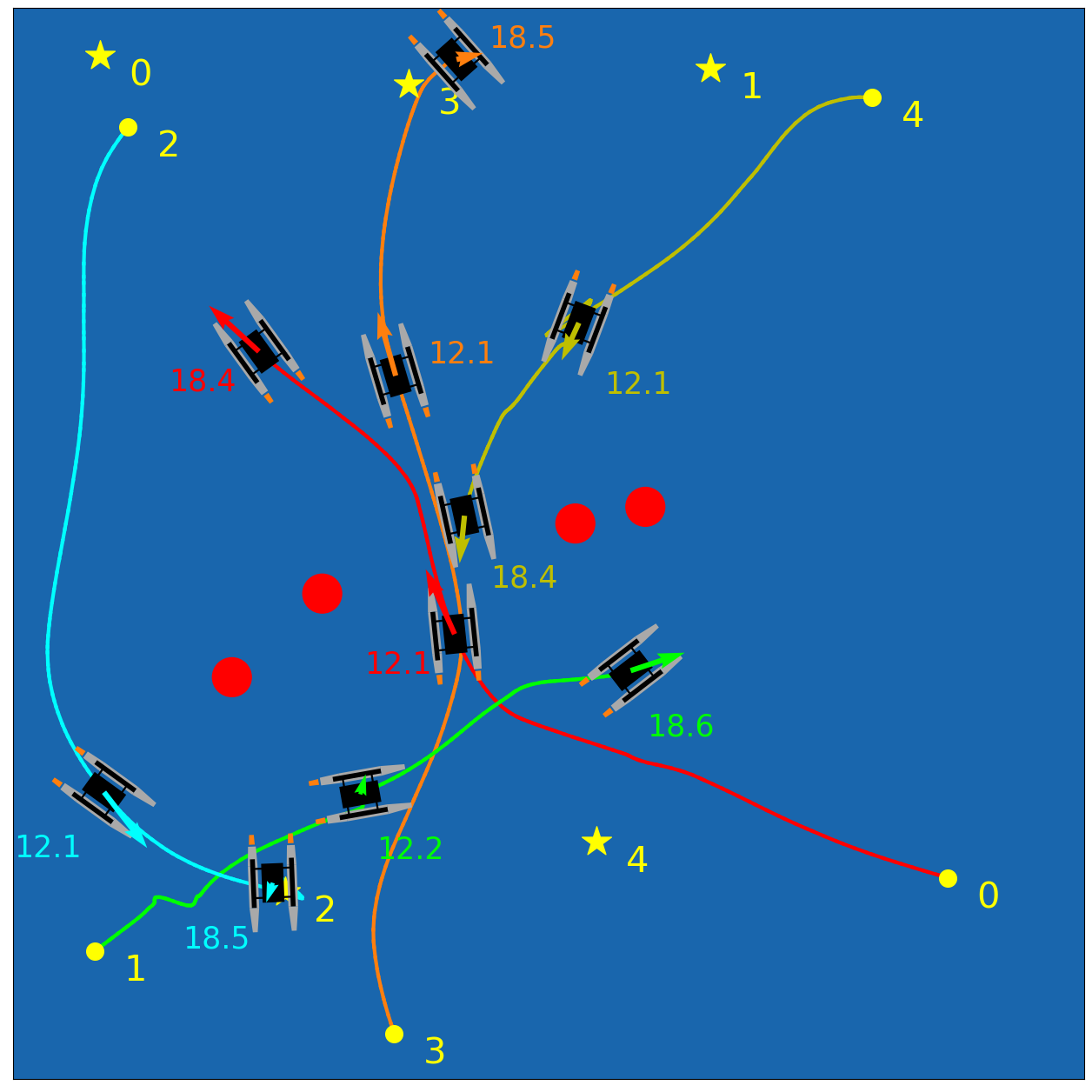}
    \end{subfigure}
    \begin{subfigure}{0.3\textwidth}
        \centering
        \includegraphics[width=\linewidth]{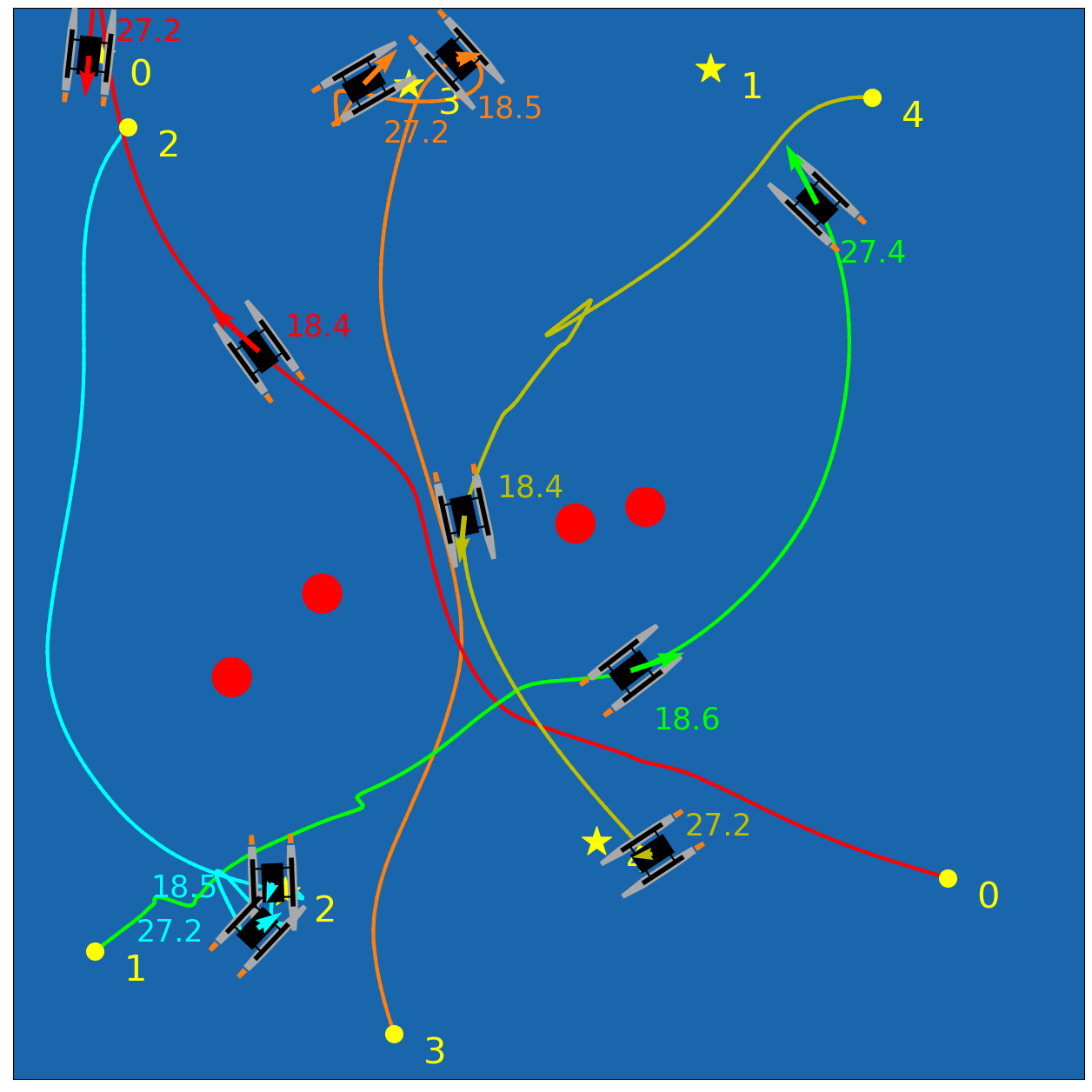}
    \end{subfigure}
    \caption{\textbf{Trajectories of AC-IQN based system in VRX experiments shown in Fig. \ref{fig:vrx exp setup}.} Yellow dots and stars indicate the start and goal positions of the vehicles, and the indices next to them match those in Fig. \ref{fig:vrx exp setup}. Vehicle poses are depicted with the vehicle icons shown. Vehicle velocities and timestamps (in seconds) are shown as arrows and numbers with one decimal place in corresponding colors. 
    }
    \label{fig:trajectories}
    \vspace{-6mm}
\end{figure*}
\begin{figure}
    \centering
    \includegraphics[width=\linewidth]{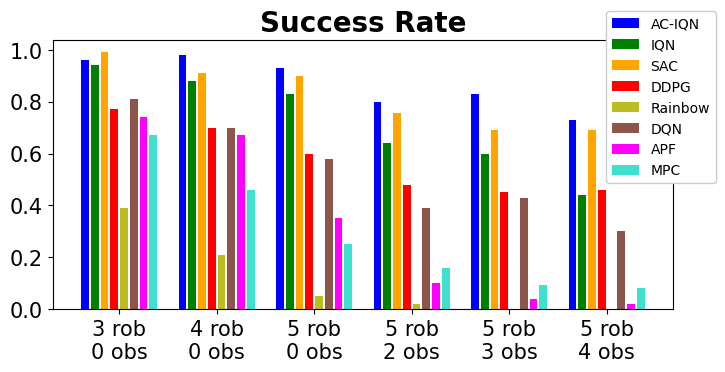}
    \caption{\textbf{Success rates in VRX experiments.} Labels on the $x$-axis indicate the number of vehicles (rob) and buoys (obs) used in corresponding sets of experiments.}
    \label{fig:exp success rate}
\end{figure}

The MPC agent we have implemented is similar to \cite{hagen2018mpc}.
The idea is to predict scenarios $k\in\{1,2,\dots,N_s\}$ corresponding to different motion commands and states of objects $i\in\{1,2,\dots,N_o\}$ in the future horizon from $t_0$, and choose the control behavior $k^*(t_0) = \arg\min_k \mathcal{H}^k(t_0)$, 
where $\mathcal{H}^k(t_0) = \max_i \max_{t\in\mathcal{D}(t_0)}(\mathcal{C}_i^k(t)\mathcal{R}_i^k(t)+\kappa_i\mathcal{M}_i^k(t)+\lambda_i\mathcal{T}_i^k(t)) +f(u_m^k,\mathcal{X}_m^k).$
As defined in \cite{hagen2018mpc}, $\mathcal{C}_i^k(t)$ are $\mathcal{R}_i^k(t)$ are collision cost and collision risk factor, $\mathcal{M}_i^k(t)$ and $\kappa_i$ are the COLREGs violation cost and a tuning parameter, $\lambda_i\mathcal{T}_i^k(t))$ is the COLREGs-transitional cost, and $f(u_m^k,\mathcal{X}_m^k)$ is the cost of maneuvering effort.
For each combination of actions from the action set $\mathbf{A}$ in Sec. \ref{subsec:action commands}, we simulate the scenario for a horizon of $T=5$ control time steps, and choose the one the minimizes the cost $\mathcal{H}^k(t_0)$.
We set $\mathcal{C}_i^k(t)=50.0$, $\kappa_i=1.0$, $\mathcal{M}_i^k(t)=10.0$, and $\lambda_i=2.0$ when the situations corresponding to the cost terms happen, otherwise these values are set to zero.
Other terms remain the same as defined in \cite{hagen2018mpc}.

Our experimental results are shown in Figure \ref{fig:exp success rate} and Table \ref{tab:exp travel time}.
In general, the AC-IQN based system achieves the highest success rate, with minimal average travel time.
APF and MPC based systems exhibit worse safety performance than RL based systems, especially in complex cases. RL agents have been trained to be robust to the influences of environmental disturbances on ASV motion and perception, and learn safety-preserving behaviors during the training process.
Among RL agents, the SAC based system demonstrates the most competitive safety performance with AC-IQN throughout all sets of experiments, but SAC requires over 40\% more travel time on average than the latter.
Accordingly, AC-IQN offers superior performance in complex multi-ASV navigation scenarios by combining continuous space maneuverability and the efficacy of Distributional RL in highly uncertain environments.
\begin{table}
    \centering
    
    \begin{tabular}{>{\centering\arraybackslash}m{1.3cm}>{\centering\arraybackslash}m{0.7cm}>{\centering\arraybackslash}m{0.7cm}>{\centering\arraybackslash}m{0.7cm}>{\centering\arraybackslash}m{0.7cm}>{\centering\arraybackslash}m{0.7cm}>{\centering\arraybackslash}m{0.7cm}}
         \toprule
         Avg. travel & 3 rob & 4 rob & 5 rob & 5 rob & 5 rob & 5 rob  \\ time (s) & 0 obs & 0 obs & 0 obs & 2 obs & 3 obs & 4 obs\\ \midrule
        AC-IQN & 17.22 & 17.85 & 20.22 & 21.06 & 21.97 & 22.81 \\ [0.5ex]
        IQN & 29.19 & 30.06 & 36.77 & 36.65 & 41.21 & 39.64 \\ [0.5ex]
        SAC & 24.09 & 24.75 & 29.96 & 29.50 & 31.54 & 32.35 \\ [0.5ex]
        DDPG & 18.22 & 18.24 & 20.51 & 20.89 & 21.34 & 21.58 \\ [0.5ex]
        Rainbow & 36.46 & 36.56 & 43.33 & 47.32 & -- & -- \\ [0.5ex]
        DQN & 19.39 & 23.38 & 21.91 & 22.24 & 23.40 & 23.25 \\
        [0.5ex]
        APF & 22.37 & 23.07 & 25.76 & 23.82 & 24.62 & 24.53 \\
        [0.5ex]
        MPC & 17.96 & 19.10 & 20.71 & 28.80 & 33.16 & 33.61 \\
        \bottomrule
    \end{tabular}
    \caption{\textbf{Travel time data from VRX experiments.} Rainbow based system fails completely in sets \textit{5 rob 3 obs} and \textit{5 rob 4 obs.}}
    \label{tab:exp travel time}
\end{table}

The initial configurations of two example experiment episodes are shown in Fig. \ref{fig:vrx exp setup}, and the behaviors of our AC-IQN based system are demonstrated in Fig. \ref{fig:trajectories}.
In each plot of Fig. \ref{fig:trajectories}, we show vehicle poses and velocities at two timestamps to clarify the relative movements.
There are slight differences in timestamps because trajectories and corresponding timestamps are recorded by vehicles individually, and we visualize those with the smallest time differences.  

As shown in the plots on the top row of Fig. \ref{fig:trajectories}, five vehicles are navigating with no buoys present in the environment.
Vehicles 0 and 1 approach each other in a head-on manner, and both decide to turn right at around 5.0 seconds to avoid a potential collision.
Vehicle 4 gets close to vehicle 3 from the port side of the latter, then it stays and gives way to vehicle 3 from around 8.0 seconds to 12.5 seconds.
After vehicle 3 passes, vehicle 4 approaches another vehicle, vehicle 1, from the port side of the other again, and it slows down to wait for clearance.
Therefore, the AC-IQN based system shows the capability of generating actions that follow COLREGs.

At bottom of Fig. \ref{fig:trajectories}, four buoys exist and form a passage in the middle.
Due to the existence of buoys, the motions of vehicles are constrained and it would be difficult and even unsafe to follow COLREGs.    
It can be seen from the plots on the bottom row of Fig. \ref{fig:trajectories} that vehicles 3, 0, and 1 approach the passage from the lower side in sequence, and vehicle 4 approaches from the upper side.
The vehicles pass in a first-in-first-out manner, where the one being closest to the passage gets to cross while others slow down and give way.  
Being the last one waiting in the line, vehicle 1 decides to detour to save time.
This episode shows that when facing difficulties in following COLREGs in a complex environment, the AC-IQN based system can maneuver in a way that is beneficial to navigation safety and efficiency. 

\section{Conclusion and Future Work}
We propose a novel ASV autonomous navigation system that integrates a perception module that works with onboard LiDAR and odometry sensors, and a decision making and control module using Distributional RL that can generate arbitrary control commands in continuous action space.
The proposed system is extensively evaluated in realistic Gazebo simulations against seven baseline systems based on state-of-the-art Distributional RL, non-Distributional RL and classical methods, and demonstrates superior performance in navigation safety and efficiency over baseline systems.  
In future work, our system performance may be further improved by introducing risk sensitivity, 
and by conducting real-world ASV field tests 
with onboard LiDAR, GPS and IMU.

\label{sec:conclusion}


\bibliographystyle{IEEEtran}
\bibliography{main}

\end{document}